\documentclass[letterpaper, 10 pt, conference]{ieeeconf}

\IEEEoverridecommandlockouts

\usepackage{amsmath}
\usepackage{amssymb}
\usepackage{bm}

\usepackage{graphics}
\usepackage{wrapfig}

\usepackage[tight]{subfigure}

\usepackage{tabularx}
\usepackage{booktabs}
\usepackage{multirow}
\usepackage{xspace}
\usepackage{algorithm}
\usepackage{algorithmic}

\usepackage{tikz}
\usetikzlibrary{decorations.text}

\usepackage[ancient]{flushend}

\usepackage{color}

\usepackage{hyperref}
\usepackage{url}
\usepackage{ulem}
\makeatletter
\let\NAT@parse\undefined
\makeatother

\usepackage[numbers,sort&compress]{natbib}
\bibpunct{[}{]}{,}{n}{}{;}

\DeclareMathOperator*{\argmax}{arg\,max}

\title{Inferring Compact Representations for\\ Efficient Natural Language Understanding of Robot Instructions}

\author {Siddharth Patki \hspace{20pt} Andrea F.\ Daniele \hspace{20pt} Matthew R.\ Walter \hspace{20pt} Thomas M.\ Howard
\thanks{Siddharth Patki and Thomas M.\ Howard are with the University of Rochester, Rochester, NY USA, {\tt\small spatki@ur.rochester.edu, thomas.howard@rochester.edu}}
\thanks{Andrea F.\ Daniele and Matthew R.\ Walter are with the Toyota Technological Institute at Chicago, Chicago, IL USA, {\tt\small \{afdaniele,mwalter\}@ttic.edu}}
}

\begin{document}
\maketitle

\begin{abstract}
    The speed and accuracy with which robots are able to interpret natural language is fundamental to realizing effective human-robot interaction. A great deal of attention has been paid to developing models and approximate inference algorithms that improve the efficiency of language understanding. However, existing methods still attempt to reason over a representation of the environment that is flat and unnecessarily detailed, which limits scalability. An open problem is then to develop methods capable of producing the most compact environment model sufficient for accurate and efficient natural language understanding. We propose a model that leverages environment-related information encoded within instructions
    to identify the subset of observations and perceptual classifiers necessary to perceive a succinct, instruction-specific environment representation. The framework uses three probabilistic graphical models trained from a corpus of annotated instructions to infer salient scene semantics, perceptual classifiers, and grounded symbols. Experimental results on two robots operating in different environments demonstrate that by exploiting the content and the structure of the instructions, our method learns compact environment representations that significantly improve the efficiency of natural language symbol grounding.
\end{abstract}

\section{Introduction} \label{sec:intro}

\begin{figure}[!t]
	\centering
	\subfigure[a mobile robot receiving a natural language instruction]{\includegraphics[width=0.95\linewidth]{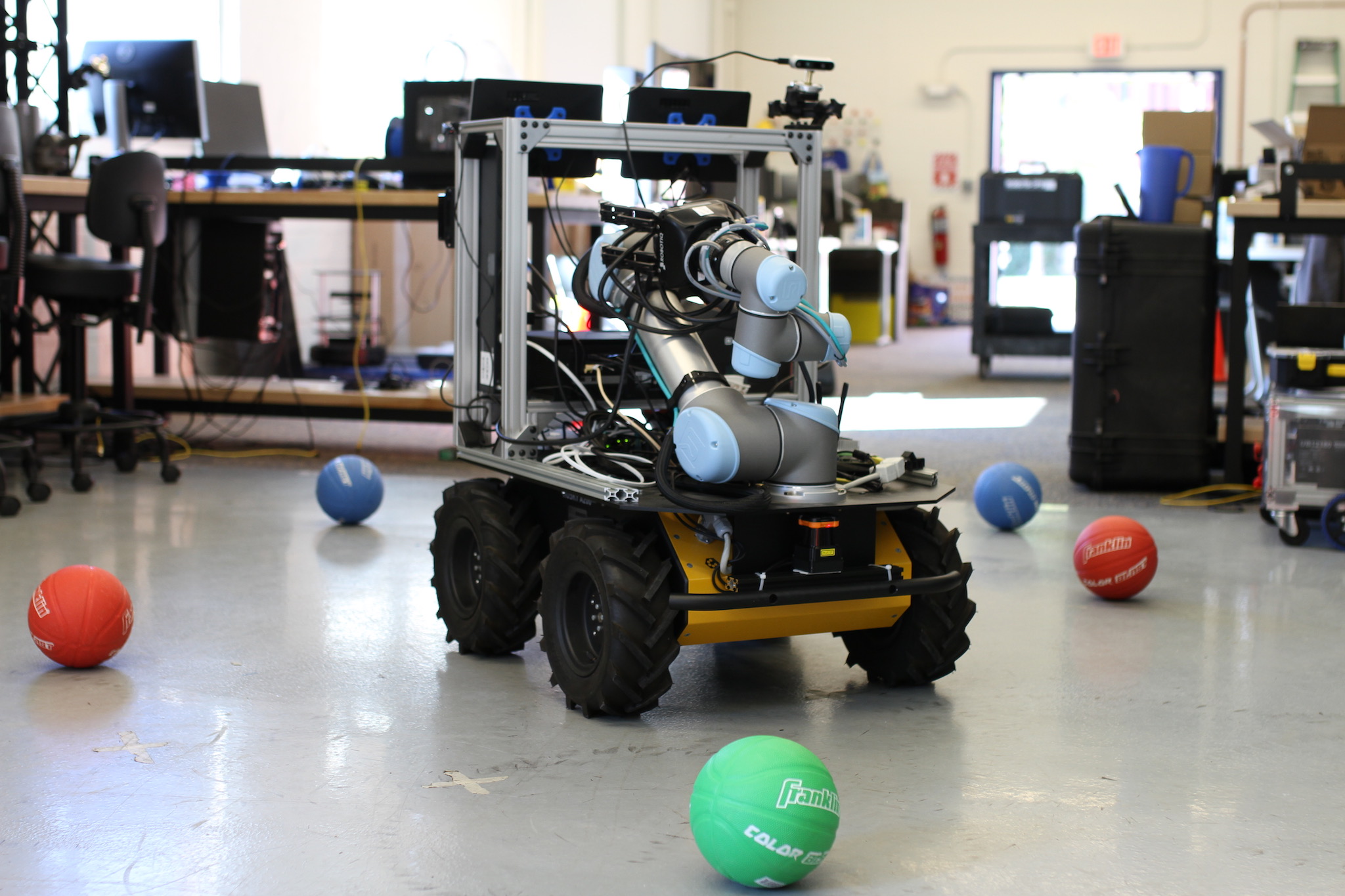}}
	\mbox{
	\subfigure[detailed world model]{\includegraphics[width=0.47\linewidth]{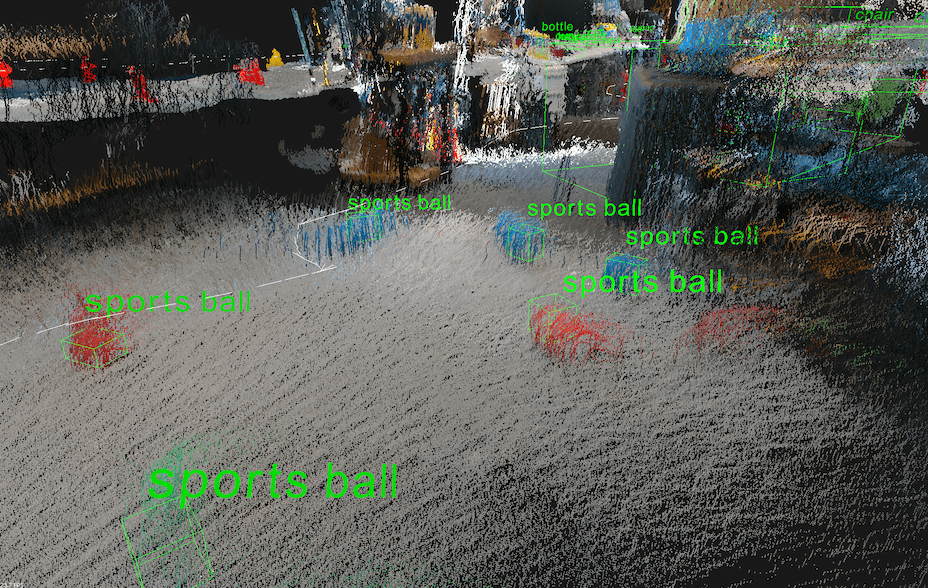}}
		\subfigure[compact world model]{\includegraphics[width=0.47\linewidth]{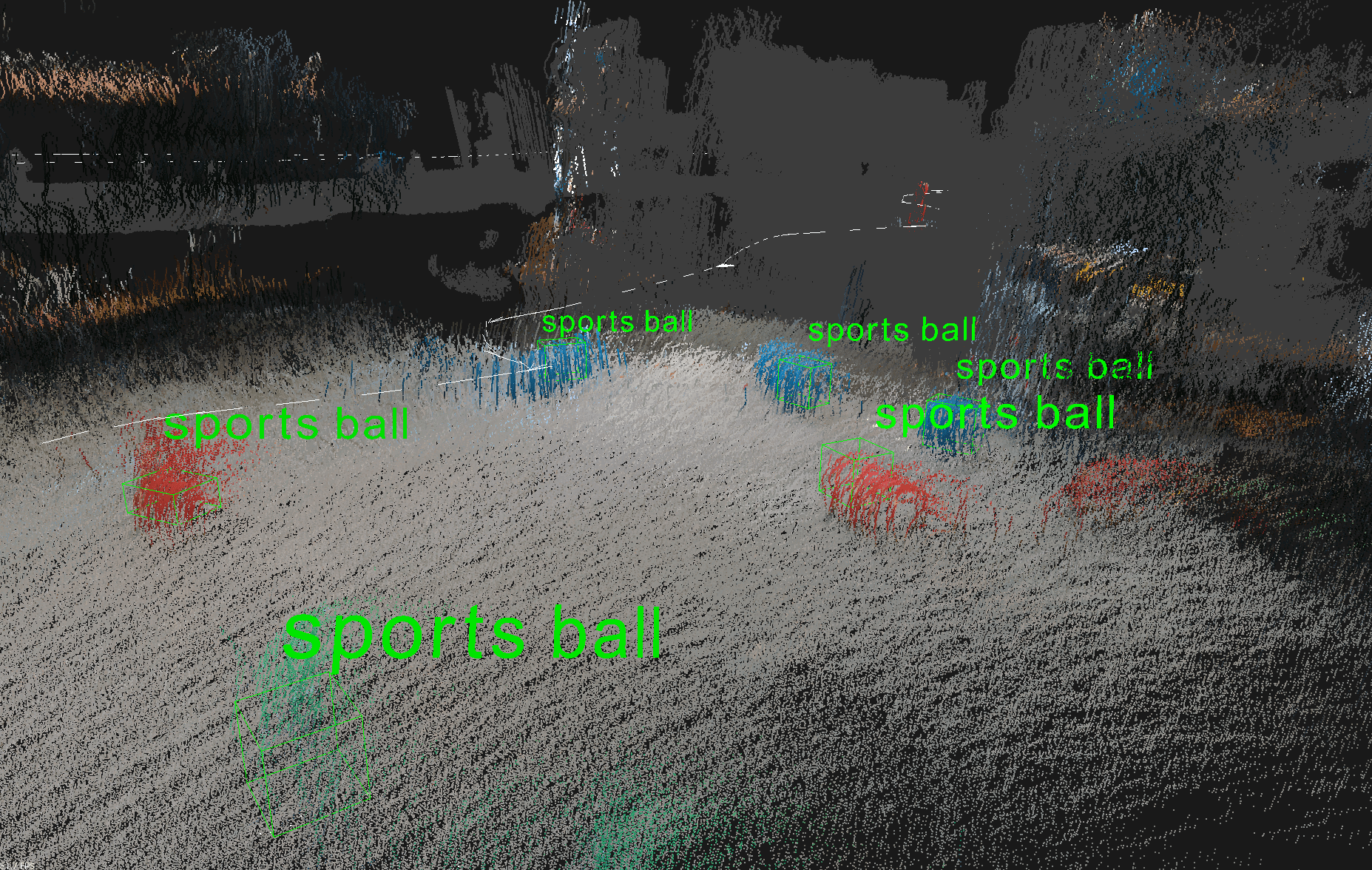}}
	}
	\caption{Our framework learns to build a minimal representation of the environment sufficient to interpret a given natural language instruction. In this example, (a) a mobile robot is directed to ``navigate to the nearest ball in the lab.'' Traditional methods interpret the instruction in the context of (b) an exhaustive world model, whereas our method maintains (c) a compact world model sufficient to ground the provided instruction.}
	\label{fig:motivation}
\end{figure}

The ability for robots to perform complex tasks is inherently linked to the richness of their environment models. Advances in sensor technology, machine perception, and natural language understanding provide a wealth of data that can be infused into these models.  These innovations raise new questions with regards to how to assimilate, manage, and utilize this abundance of knowledge. A fundamental problem is how to reason over this rich information in a manner that enables robots to efficiently plan in diverse environments of varying scales and complexities.  Consider the human-robot teaming scenario illustrated in Figure~\ref{fig:motivation}, in which a user instructs the mobile robot to ``navigate to the nearest red ball.''
If we assume that the robot has access to knowledge bases (e.g., campus-level maps) and various sensor measurements (e.g., images, laser scans, audio, etc.) that it has accumulated over time, the problem becomes one of situating or ``grounding'' the instruction in the context of the perceived environment.  With a few exceptions~\cite{kuipers04, modayil04, beeson10, pronobis12, walter13}, contemporary methods  attempt to fuse the knowledge bases and sensor measurements into a single, flat representation of the environment (i.e., the ``world model'') that expresses all metric~\cite{eustice05, olson06, durrant-whyte06, bailey06, walter07, kaess08, cummins09} as well as semantic~\cite{mozos07, zender08, pronobis12, walter13, hemachandra14, hemachandra15} knowledge gleaned from the observations. There are three fundamental limitations to this approach.

First, a consistent, high fidelity model of the environment is expensive to maintain in terms of both compute and memory storage. Second, searching over dense models is computationally prohibitive in the context of both planning and natural language understanding~\cite{tellex11a, howard14, chung15}, with costs as high as exponential in the size of the model~\cite{tellex11a}. More generally, it is unnecessarily detailed for most tasks. Ideally, one would reason over the most compact representation of the environment necessary to understand the instruction. However, this representation can not be inferred until after the instruction is received. Third, in situations in which concepts are taught or evolve in-situ from human demonstrations, previous interpretations of the environment may become incorrect or deficient, necessitating a means of revisiting these models as needed.

We propose a framework that explicitly reasons over the relevance of the observations and perceptual classifiers available, so as to learn a task-relevant, scalable environment representation sufficient for planning and natural language understanding. Underlying this method is a learned probabilistic model that can be readily adapted based upon the difficulty of the task and the complexity of the environment. Importantly, the method infers an efficient environment representation online by leveraging a learned model of saliency. This model extracts characteristics of the representation from free-form utterances to ``lazily'' reason over the small subset of available knowledge pertinent to the task. Specifically, we build upon recent work on adapting perception pipelines from natural language instructions~\cite{patki18a} to infer subsets of observations that we use to construct instruction-specific representations of the environment. These induced representations are more efficient to search, yet still express the correct hierarchies and affordances necessary to perform the task.  In scenarios where humans can interactively teach robots to classify objects in-situ, past observations of such objects could be added to the world model given utterances that reference the object.

The central contribution of this paper is a framework that exploits three probabilistic graphical models in the form of
Distributed Correspondence Graphs~\cite{howard14} to adaptively model the environment representation in a
task-specific manner. These models are trained from examples of how language maps to the relevant scene semantics, perceptual classifiers, and the symbols used to ground language-based instructions.
Experimental results demonstrate that the ability to dynamically adapt perception and observation models significantly improves the computational efficiency of natural language symbol grounding.

\section{Related Work} \label{sec:related}

Existing language understanding methods reason over a flat, unified symbolic model of the world that expresses the spatial,
semantic, and/or topologic properties of the environment through a representation that is assumed to be globally
consistent.
In practice, these models are typically constructed by running a state-of-the-art SLAM algorithm~\cite{walter07,
eustice05, olson06, kaess08, grisetti09}, which provides flat, globally metric models of the environment that are
limited to spatial information. Semantic and topologic properties are then manually injected to realize a representation
suitable for language grounding.
Localization and mapping methods that attempt to jointly reason over spatial, semantic, and topologic
properties of the environment have also been proposed~\cite{kuipers00, zender08, vasudevan08, pronobis12,
walter13, hemachandra14, duvallet14, hemachandra15}.
With few exceptions~\cite{kuipers00}, however, these methods still attempt to maintain a single globally consistent environment
representation, which is both unnecessarily detailed for language grounding and also resource (e.g., memory) intensive.

Given a natural language utterance, grounding methods~\cite{harnad90,tellex11,howard14} attempt to associate each word in the utterance with its corresponding referent in this environment model and the robot's symbolic action space. Semantic parsing-based methods~\cite{matuszek10,matuszek12a,thomason15} similarly map natural language to meaning representations, typically in the form of a lambda calculus. Early work in grounding~\cite{winograd71,roy03} employs manually engineered correspondences and features between words in a flat representation of the environment. Modern day methods~\cite{kollar10,tellex11a,tellex11,howard14,chung15} take a statistical approach to language grounding (and similarly for inverse grounding~\cite{tellex12,tellex14,gong18}) that employs probabilistic models that relate words to their corresponding referents according to the hierarchical structure of language, enabling the resolution of complex free-form language. These models are typically learned from annotated natural language corpora as well as through interaction with humans~\cite{thomason15,spranger15,she17}. Probabilistic grounding models have been shown to be effective at interpreting cooking instructions~\cite{bollini10}, learning spatial relations in semantic maps \cite{walter13,hemachandra14}, and directing mobile manipulators~\cite{walter14b}, among others.

These methods perform inference over the entire set of state and action symbols, resulting in a computational complexity that is proportional to the power set of objects, regions, and constraints. This limits inference to simple tasks with a few interchangeable constraints or requires access to a set of predefined environment-specific behaviors. To improve scalability, \citet{howard14} developed the Distributed Correspondence Graph (DCG) model that separates inference across conditionally independent constituents of the graph. In effect, this distributes inference across multiple factors in a graphical model, transforming the computational complexity from exponential to linear in the number of symbols. \citet{chung15} propose the Hierarchical Distributed Correspondence Graph (HDCG), which improves the efficiency of inference by learning to construct a more efficient approximation of the space of relevant symbols for probabilistic language grounding. \citet{paul16a} describe a method that partitions the joint distribution into concrete and abstract factors. The algortihm performs inference in two stages per phrase.  In the first stage, distributions of concrete symbols are inferred and used to inform sparse approximations of the abstract symbolic representation that are more efficient to search.  In the second stage, distributions of abstract symbols are inferred and joined with the concrete symbols to represent the meaning of each phrase.

\section{Technical Approach} \label{sec:approach}

The problem of natural language understanding is commonly framed as inference over a learned distribution that associates linguistic elements with their corresponding symbolic representation of the robot's state and action spaces. More specifically, inference involves reasoning over a representation $\Gamma_{s}$ that symbolizes objects, places, constraints, actions, trajectories, and others concepts expressed by the robot's world model. The set of symbols forms a discrete and finite space in which the instruction can be grounded. The distribution over groundings is conditioned over a parse of the utterance $\bm{\Lambda}$ as well as a world model $\bm{\Upsilon}_{t}$ expressing environment knowledge that may be known a priori $\bm{\Upsilon}_{0}$ or extracted from multimodal observations $\mathbf{z}_{1:t}$ using the classifiers in the robot's perception pipeline $\mathbf{P}$
\begin{equation}
    \bm{\Upsilon}_{t} \approx f( \mathbf{z}_{1:t}, \mathbf{P}, \bm{\Upsilon}_{0} ).
\end{equation}
Natural language understanding then follows as maximum a posteriori (MAP) inference over $\Gamma_s$
\begin{equation}
    \Gamma_{s}^* = \argmax_{ \gamma_1 ... \gamma_n \in \bm{\Gamma_{s}} }  \; p \left( \Gamma_{s}  \vert  \bm{\Lambda}, \bm{\Upsilon}_t \right).\label{eqn:basic-2}
\end{equation}
Several contemporary approaches~\cite{tellex11a,
howard14, paul16a} formulate this problem as probabilistic inference in a factor graph with a hierarchical structure
dictated by the compositional nature of the utterance, symbolic representation, and environment. This enables the model
to reason about the meaning of particular phrases in terms of the symbolic grounding space based upon their child phrases, and a model of the environment. The parameters of the grounding model (e.g., feature weights in a log-linear model) are learned from annotated corpora that express the meaning of each phrase in the context of the child groundings and phrases.

In practical settings, the the space of groundings $\Gamma_s$, the environment  $\bm{\Upsilon}_t$ is complex, and the free-form instructions $\Lambda$ may be complex and diverse, making exact inference computationally intractable. To address this, the Distributed Correspondence Graph~\cite{howard14} proposes an approximate factorization of the grounding distribution that affords an efficient inference
\begin{equation}
    \Phi_{s}^* = \argmax_{ \phi_{ij} \in \Phi_{s} }  \; \prod\limits_{ i = 1 }^{ \lvert \Lambda \rvert } \prod\limits_{ j = 1 }^{ \lvert \Gamma_{s} \rvert }  p( \phi_{ij}  \vert  \gamma_{ij}, \lambda_i, \Phi_{ci}, \bm{\Upsilon}_{t} ).
    \label{eqn:dcg_0}
\end{equation}
Formally, DCG inference involves searching for the most likely assignment of boolean correspondence variables $\Phi^*_{s}$~\cite{paul2018efficientplatforms} in the context of the groundings $\gamma_{ij} \in \Gamma_s$, phrases $\lambda_i \in \Lambda$, child correspondences $\Phi_{ci}$, and the world model $\bm{\Upsilon}_{t}$ by maximizing the factorization in Equation~\ref{eqn:dcg_0}. In such model, a correspondence variable $\phi_{ij}$ being true expresses the fact that the corresponding grounding $\gamma_{ij}$ matches the associated phrase in the command.

The ability to ground free-form instructions is inherently linked to the richness of the robot's environment
representation $\bm{\Upsilon}_{t}$. However, building exhaustively detailed world models using all available knowledge
bases and observations $\mathbf{z}_{1:t}$ is computationally expensive, particularly in large-scale, unstructured
environments. The runtime of common language understanding models such as $\text{G}^3$ are exponential in the cardinality of the symbol space $\lvert \Gamma_{s}\rvert$~\cite{howard14}. DCG improves this complexity to being linear in the size of the world model, however the cost of inference still inhibits real-time human-robot interaction.

In practice, a large fraction of the objects and their corresponding symbols that comprise the inferred world model
are typically inconsequential to the meaning of the utterance. In such cases, there exists a compact environment representation $\bm{\Upsilon}_{t}^{*}$ that is sufficient to interpret the utterance, providing a significant improvement in the computational efficiency of inference
relative to the standard model (Equation~\ref{eqn:dcg_0}).

We propose a probabilistic model that exploits natural language in order to guide the generation of these compact world
models $\bm{\Upsilon}_{t}^{*}$. Integral to this approach is the ability to infer a small, succinct subset of perceptual
classifiers $\mathbf{P}^{*} \in \mathbf{P}$ in a manner that dynamically adapts the robot's perceptual capabilities
according to the current task
\begin{equation}
    \mathbf{P}^{*} \approx f\left( \mathbf{P}, \bm{\Lambda} \right), \label{eqn:compact_P}
\end{equation}
resulting in the compact world model
\begin{equation}
    \bm{\Upsilon_{t}}^{*} \approx f\left( \mathbf{z}_{1:t}, \mathbf{P}^{*}, \bm{\Upsilon}_{0} \right)
\end{equation}

We further observe that not all observations are necessary to produce this compact representation $\bm{\Upsilon}_{t}^{*}$.  For instructions in which the context of the observation may be evident (e.g., ``drive to the nearest red ball in the hallway''), samples outside of these semantically classified regions (i.e., hallways) can be pruned from the space of observations.
As the robot drives through the environment, a real-time scene classifier produces a semantic label (i.e., a scene category) that will be associated with all of the observations (from all available sensors) and pose measurements. The ability to assign a label to the current region in real-time allows us to treat such information as an observation produced by a virtual sensor (i.e., the scene classifier).

We define a minimal set of observations $\mathbf{z}^{*} \in \mathbf{z}_{1:t}$ that, based on their semantic labels, are used to construct the compact representation that is sufficiently detailed to contain all symbols necessary to be expressed
by the natural language symbol grounding model
\begin{subequations}
    \begin{align}
        \mathbf{z}^{*} &\approx f\left( \mathbf{z}_{1:t}, \bm{\Lambda} \right) \label{eqn:compact_z}\\
        \bm{\Upsilon_{t}}^{*} &\approx f\left( \mathbf{z}^{*}, \mathbf{P}^{*}, \bm{\Upsilon}_{0} \right). \label{eqn:compact_w}
    \end{align}
\end{subequations}

Using the subsampled set of observations to construct a compact representation for symbol grounding transforms the expression for natural language inference (Eqn.~\ref{eqn:dcg_0}) to
\begin{equation}
    \Phi_{s}^* = \argmax_{ \phi_{ij} \in \Phi_{s} }  \; \prod\limits_{ i = 1 }^{ \lvert \Lambda \rvert } \prod\limits_{ j = 1 }^{ \lvert \Gamma_{s} \rvert }  p( \phi_{ij}  \vert  \gamma_{ij}, \lambda_i, \Phi_{ci}, \bm{\Upsilon_{t}^{*}} ).
\label{eqn:fullmodel}
\end{equation}

This inference problem
requires that we learn three models (Fig.~\ref{fig:architecture}): an adaptive perception model, an observation filtering model, and a natural language
symbol grounding model.
The process for training these models begins with the natural language symbol grounding module, in which symbols that
represent objects, spatial relationships, containers, constraints, actions, and other types are associated with language
~\cite{howard14, paul16a}. The process of training the observation filtering and adaptive perception models requires
one to fit the minimum set of semantic labels and perceptual classifiers.
Such classifiers are the ones that extract the most compact environment representation for each example that will
not prune out any of the annotated ground-truth symbols from the corpus of instructions.
This process yields three separate corpora with common instructions, but different symbolic representations and
annotations that we use to train the three distinct models.

\section{Experimental Setup}
\label{sec:experiments}
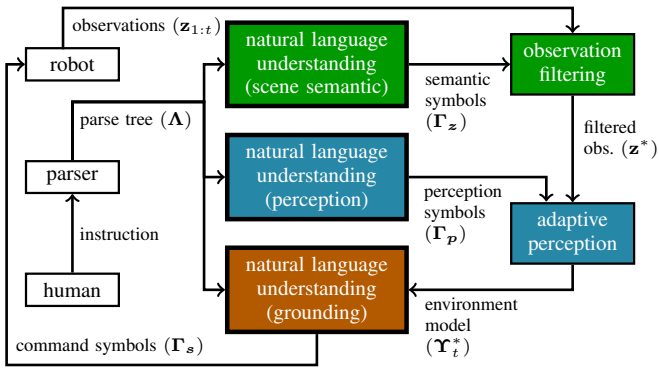
\begin{figure}[!t]
\centering
\vspace{6pt}
\begin{tikzpicture}
\node[rectangle,font=\footnotesize,draw=black, text=white, fill=black!40!cyan, line width=0.03cm, text width=1.4cm, align=center] at (3.65,-0.75) (ap) {adaptive perception};
\node[rectangle,font=\footnotesize,draw=black, line width=0.03cm, text width=1cm, align=center] at (-3.0,0) (parser) {parser};
\node[rectangle,font=\footnotesize,draw=black, text=white, fill=black!40!green, line width=0.03cm, text width=1.4cm, align=center] at (3.65,1.5) (sc) {observation filtering};
\node[rectangle,font=\footnotesize,draw=black, line width=0.03cm, text width=1cm, align=center] at (-3.0,1.5) (robot) {robot};
\node[rectangle,font=\footnotesize,draw=black, line width=0.03cm, text width=1cm, align=center] at (-3.0,-1.5) (human) {human};
\node[rectangle,font=\footnotesize,draw=black, text=white, fill=black!40!green, line width=0.06cm, text width=2.15cm, align=center] at (0.25,1.5) (nlsgsem) {natural language understanding\\ (scene semantic)};
\node[rectangle,font=\footnotesize,draw=black, text=white, fill=black!40!cyan, line width=0.06cm, text width=2.15cm, align=center] at (0.25,0) (nlsgper) {natural language understanding (perception)};
\node[rectangle,font=\footnotesize,draw=black, text=white, fill=black!30!orange, line width=0.06cm, text width=2.15cm, align=center] at (0.25,-1.5) (nlsggrd) {natural language understanding (grounding)};
\draw[->,line width=1pt] (nlsgsem) to (sc);
\draw[->,line width=1pt] (sc) to (ap);
\draw[->,line width=1pt] (human) to (parser);
\draw[->,line width=1pt] (parser) to (-3.0,1) to (-1.25,1) to (-1.25,-1.5) to (nlsggrd);
\draw[->,line width=1pt] (parser) to (-3.0,1) to (-1.25,1) to (-1.25,0) to (nlsgper);
\draw[->,line width=1pt] (parser) to (-3.0,1) to (-1.25,1) to (-1.25,1.5) to (nlsgsem);
\draw[->,line width=1pt] (-3.125,1.7) to (-3.125,2.25) to (3.65,2.25) to (sc);
\draw[->,line width=1pt] (nlsgper) to (3.3,0) to (3.3,-0.325);
\draw[->,line width=1pt] (ap) to (3.65,-1.5) to (nlsggrd);
\draw[->,line width=1pt] (nlsggrd) to (0.25,-2.5) to (-3.875,-2.5) to (-3.875,1.5) to (robot);
\node[font=\scriptsize,text width=2cm,align=center] at (-2.15,0.75) (parse tree) {parse tree $\left(\bm{\Lambda}\right)$};
\node[font=\scriptsize,text width=1.5cm] at (-2.15,-0.75) (instruction) {instruction};
\node[font=\scriptsize,text width=4.5cm] at (-1.5,-2.25) (command) {command symbols $\left(\bm{\Gamma_{s}}\right)$};
\node[font=\scriptsize,text width=1.625cm,align=left] at (2.5,1.0) (semantic) {semantic\\ symbols\\ $\left(\bm{\Gamma_{z}}\right)$};
\node[font=\scriptsize,text width=1.75cm,align=left] at (2.55,-0.5) (perception) {perception\\ symbols \\$\left(\bm{\Gamma_{p}}\right)$};
\node[font=\scriptsize,text width=3cm] at (-1.5,2.0) (observations) {observations $\left(\mathbf{z}_{1:t}\right)$};
\node[font=\scriptsize,text width=1.0cm] at (4.3,0.5) (filteredobservations) {filtered obs.\ $\left(\mathbf{z}^{*}\right)$};
\node[font=\scriptsize,text width=1.625cm,align=left] at (2.5,-2.0) (adapted environment) {environment model\\ $\left(\bm{\Upsilon}_{t}^{*}\right)$};
\end{tikzpicture}
\caption{The system architecture for language-guided observation filtering, adaptive perception, and natural language symbol grounding.  The three natural language understanding models that are learned from the annotated instructions are highlighted in bold.}
\label{fig:architecture}
\end{figure}

Figure~\ref{fig:architecture} illustrates the software architecture that we implemented for experimental evaluation of the proposed algorithm.  In this architecture, the robot stores the sensors measurements in the observation filtering module.
When the human provides a textual instruction, we convert the text into a parse tree $\bm{\Lambda}$ that is provided to the three
natural language understanding modules. The \textit{scene semantics} natural language understanding module extracts the salient
scene semantics $\bm{\Gamma_{z}}$ pertaining to the instruction. The observations filtering module then extracts a subset of observations $\mathbf{z}^{*}$ (Eqn.~\ref{eqn:compact_z}) based on the inferred scene semantic label(s). The \textit{perception} natural language understanding module extracts the symbols representing the classifiers (Eqn.~\ref{eqn:compact_P}) that are necessary to detect the objects that are relevant to the natural language instruction. This information is then passed to the adaptive perception node that extracts an approximation of the environment model $\bm{\Upsilon_{t}}^{*}$ (Eqn.~\ref{eqn:compact_w}) from $\mathbf{z}^{*}$ using the sub-sampled classifiers $\mathbf{P}^{*}$. The \textit{symbol grounding} natural language understanding module uses the parse tree and the world model approximation to extract a distribution of symbols that represents the robot behavior $\bm{\Gamma_{s}}$ (Eqn.~\ref{eqn:fullmodel}).

All of the natural language understanding modules are implemented as Distributed Correspondence Graphs~\cite{howard14}
with symbolic representations and features adapted for each of the scene semantics, perception, and grounding domains.

We trained the natural language understanding modules with a synthetic corpus of annotated examples consistent with example robot instructions, such as ``navigate to the nearest cone in the parking lot'' or ``navigate to the farthest blue ball.''  Approximately 500 instructions were annotated for the scene semantic, perception, and grounding models in accordance with their symbolic representation. The software was integrated onto two Clearpath Robotics Husky A200 Unmanned Ground Vehicles (Fig.~\ref{fig:motivation}) and used for dataset collection at two distinct sites. Visual observations were collected using the RealSense D435 RGB-D sensor. Robot localization was performed using laser-scan matching with a planar LIDAR sensor.

In these experiments, we use eight semantic labels such as ``kitchen,'' ``laboratory,'' ``parking lot,'' etc., which are
associated with sensor observations. To detect the semantics of the scene ,we use a YOLO object detector~\cite{redmon2017yolo9000} trained on the COCO dataset~\cite{lin2014microsoft}. Object detections are passed to a scene classifier. The scene classifier then assigns a semantic label to each observation based on an object co-occurrence model that relates objects and scene classes. Objects that are not characteristic of any particular scene (e.g., person, cat, or horse) are ignored. The perception pipeline within in the adaptive perception node contains multiple elements including a YOLO-based object detector, a noise removal filter that refines the segmented object clusters, a 3d bounding box detector, an LUV color space-based color detector, and a 3-DOF pose detector. We limit the sensing range to 3.5\,m to avoid processing noisy point cloud data.

The experiments were designed to explore the impact of observation filtering and adaptive perception on the task of mobile robot instruction following. We quantify the performance of the system using metrics of computational efficiency of perception for
symbol grounding under the assumption of lazy evaluation of the observations.

\section{Results} \label{sec:results}
This section presents results highlighting the performance of different aspects of the learned models in our proposed architecture.  First, we highlight the computational efficiency of adaptive perception applied in the navigation domain.  Second, we demonstrate how observation filtering reduces the number of observations we need to reason over in order to extract task-relevant objects. Later, we demonstrate the efficiency gains achieved by combining these two strategies in order to generate compact world representations.

\subsection{Adaptive Perception}
\label{sec:results-adaptive-perception}

In previous experiments~\cite{patki18a}, we observed that language grounding was faster in environments inferred by adaptive perception than non-adaptive perception. Also the adaptive perception was found to be faster than its counterpart. To verify the predicted behavior of the adaptive perception pipeline, we analyzed its impact on the runtime of perception by evaluating it on the datasets collected at two diffrent sites for six different instructions. Table~\ref{table:results-1} presents the results demonstrating the impact of adaptive perception (AP) on the perception runtime against the standard baseline (B) that corresponds to the standard approach of invoking all classifiers and observations. Table~\ref{table:results-2} shows the impact of adaptive perception on the compactness of the approximated world representations. Consistent with previous evaluations~\cite{patki18a}, reducing the cardinality of the world model improves the runtime of language grounding.

\begin{figure}[!t]
	\centering
	\vspace{6pt}
	\subfigure[exhaustive perception: detecting all objects]{\includegraphics[width=1.0\linewidth]{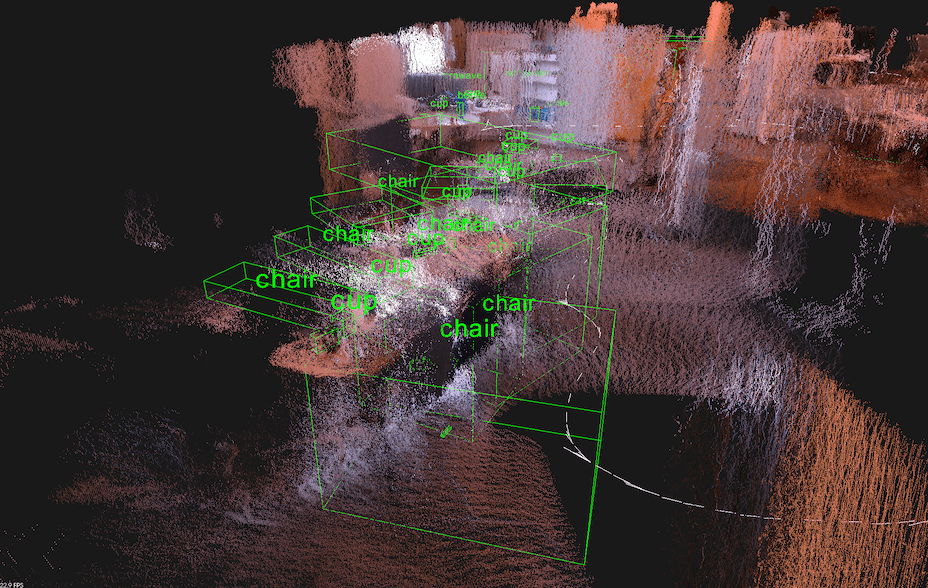}\label{fig:exhaustive}}\\
	\subfigure[adaptive perception: detecting only cups]{\includegraphics[width=1.0\linewidth]{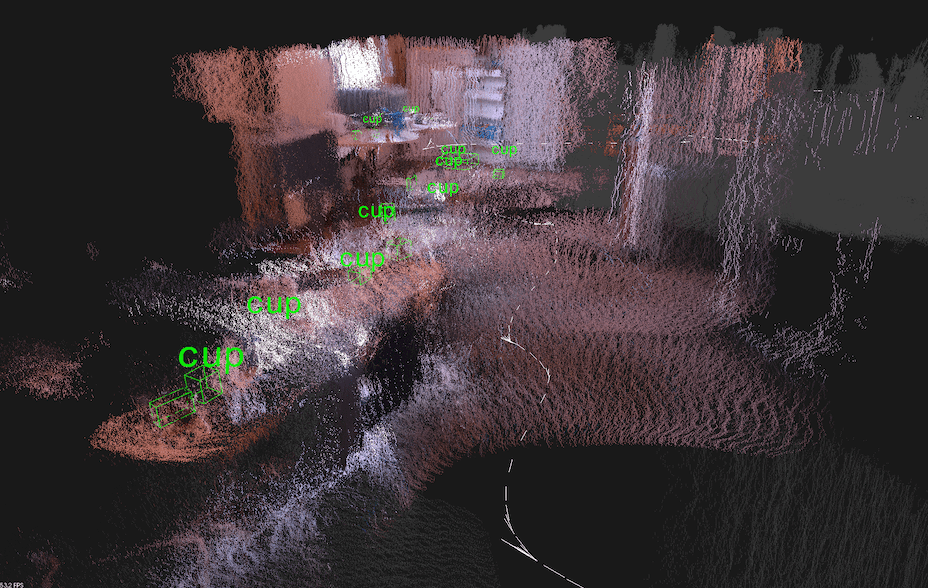}\label{fig:adaptive}}\\
	\caption{Impact of adaptive perception for the command ``drive to the farthest cup in the kitchen.'' A standard approach requires generating and reasoning over \subref{fig:exhaustive} an exhaustive map generated using all of the available object detectors, resulting in a map with $37$ objects and a runtime of $408$\,s. In contrast, our adaptive method generates \subref{fig:adaptive} a more compact map only using detectors relevant to the command, resulting in a map with $11$ objects and a runtime of $225$\,s.}.
	\label{fig:adaptive-perception-navigation-results}
\end{figure}

Figure \ref{fig:adaptive-perception-navigation-results} demonstrates the impact of adaptive perception for the example instruction ``drive to the nearest cup in the kitchen.'' In this particular example, the model is able to independently evaluate which object detectors should be engaged to construct an instruction-specific world model. By using the information contained within the instructions, our method results in a $36\%$ reduction in the time required to build an environment representation for inferring the instruction ``go to the nearest cup in the kitchen.'' This demonstrates how inferring the classifiers useful for generating task-relevant compact representations can reduce the runtime requirements of robot perception. As we have seen~\cite{patki18a}, the reduction in runtime is proportional to the sparsity of classifiers necessary to extract a sufficient detailed environment model that is suitable for the grounding of specific instructions.

As more complex detectors are considered (e.g., ICP-based point cloud matching), we expect to find that these differences will become increasingly significant.  For example, an operator performing service on a truck may require a robot to ``turn the top-left screw on the back panel by forty five degrees'' at one point during an activity, while it may also ask the same robot to ``unload the truck of all of the pallets'' at a later time.  The computational requirements of the multitude of classifiers necessary to generate a consistent interpretation of the environment that is sophisticated enough to perform both of these tasks may be too burdensome for an robot to extract in real-time. We hypothesize that as the interactions approach such diversity and complexity, a model that extracts the salient information from the command and constructs a representation suitable for natural language symbol grounding will outperform non-adaptive representations of the environment.

\subsection{Observation Filtering}
\label{sec:results-observation-filtering}
\begin{table}[!t]
	\centering
	\vspace{6pt}
	\caption{ Improvement in the perception runtime at sites 1 \& 2 }
	\label{table:results-1}
	\setlength{\tabcolsep}{4.0pt}
	{\scriptsize
	\begin{tabularx}{1.0\linewidth}{lccccc}
		\toprule
			& & \multicolumn{4}{c}{ (runtime in seconds) } \\
	  	Instruction  &  Site & B & OF & AP & OF+AP \\
	  	\midrule
	  	``go to the farthest umbrella in the hallway'' & 1 & 401 & 60 & 242 & 55 \\
	  	``go to the nearest suitcase in the parking lot'' & 2 & 306 & 136 & 220 & 99 \\
	  	``go to the farthest cup in the kitchen'' & 1 & 401 & 146 & 225 & 75 \\
	  	``go to the nearest keyboard in the office'' & 2 & 306 & 74 & 222 & 46 \\
	  	``go to the nearest ball in the hallway'' & 1 & 401 & 59 & 217 & 38 \\
			``go to the farthest ball in the lab'' & 2 & 306 & 67 & 206 & 48 \\
	  \bottomrule
  \end{tabularx}}
\end{table}
\begin{table}[!t]
	\centering
	\caption{ Improvement in the representation compactness at sites 1 \& 2  }
	\label{table:results-2}
	\setlength{\tabcolsep}{4.5pt}
	{\scriptsize
	\begin{tabularx}{1.0\linewidth}{lccccc}
		\toprule
			& & \multicolumn{4}{c}{ (\# of detected objects) } \\
	  	Instruction  &  Site & B & OF & AP & OF+AP \\
	  	\midrule
	  	``go to the farthest umbrella in the hallway'' & 1 & 37 & 4 & 2 & 2 \\
	  	``go to the nearest suitcase in the parking lot'' & 2 & 36 & 3 & 3 & 2 \\
	  	``go to the farthest cup in the kitchen'' & 1 & 37 & 29 & 11 & 9 \\
	  	``go to the nearest keyboard in the office'' & 2 & 36 & 29 & 3 & 3 \\
	  	``go to the nearest ball in the hallway'' & 1 & 37 & 4 & 1 & 1 \\
	  	``go to the farthest ball in the lab'' & 2 & 36 & 7 & 7 & 7 \\
	  \bottomrule
  \end{tabularx}}
\end{table}
To explore the impact of observation filtering, we evaluated the runtime performance of perception on the same six instructions explored for the adaptive perception experiment. Table~\ref{table:results-1} presents the results that reveal the impact of observation filtering (OF) against the standard baseline (B).  This result demonstrates how removing observations inferred to be unnecessary to extract the meaning of the natural language instruction can improve the runtime performance of robot perception.  The results demonstrate a $55\%$ reduction in runtime for the instruction ``go to the nearest suitcase in the parking lot'' over the baseline.  The improvement is a function of the diversity of scene labels across all observations. Table~\ref{table:results-2} shows the impact of observations filtering on the compactness of the approximated world representation. In this case the improvement is a function of the distribution of objects across different regions in the world.

\subsection{Observation Filtering with Adaptive Perception}
\begin{figure*}[!t]
	\centering
	\subfigure[exhaustive environment model]{\includegraphics[width=0.325\linewidth]{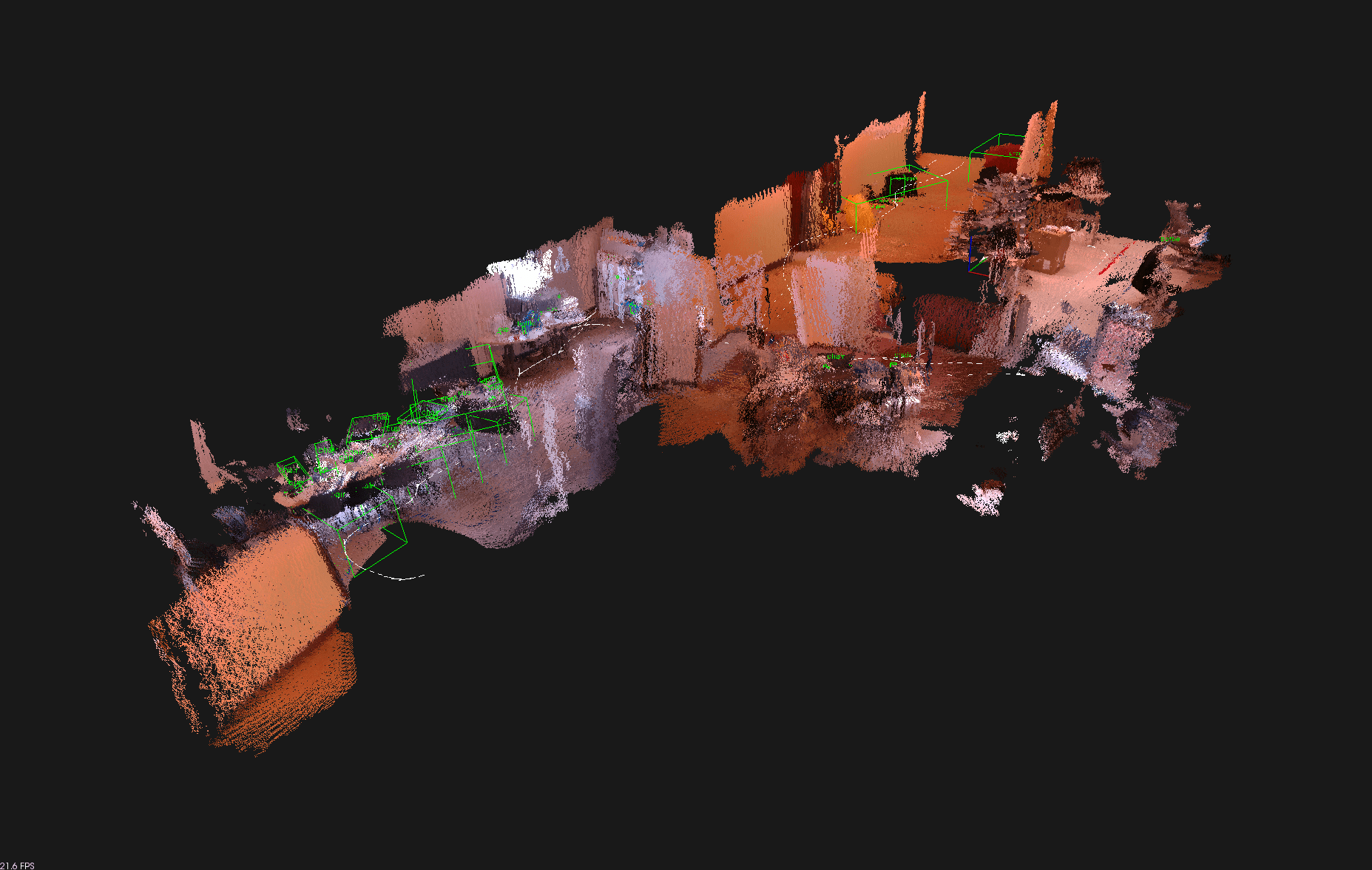}\label{fig:exhaustive-map-ttic}}
	\subfigure[semantic scene labels]{\includegraphics[width=0.325\linewidth]{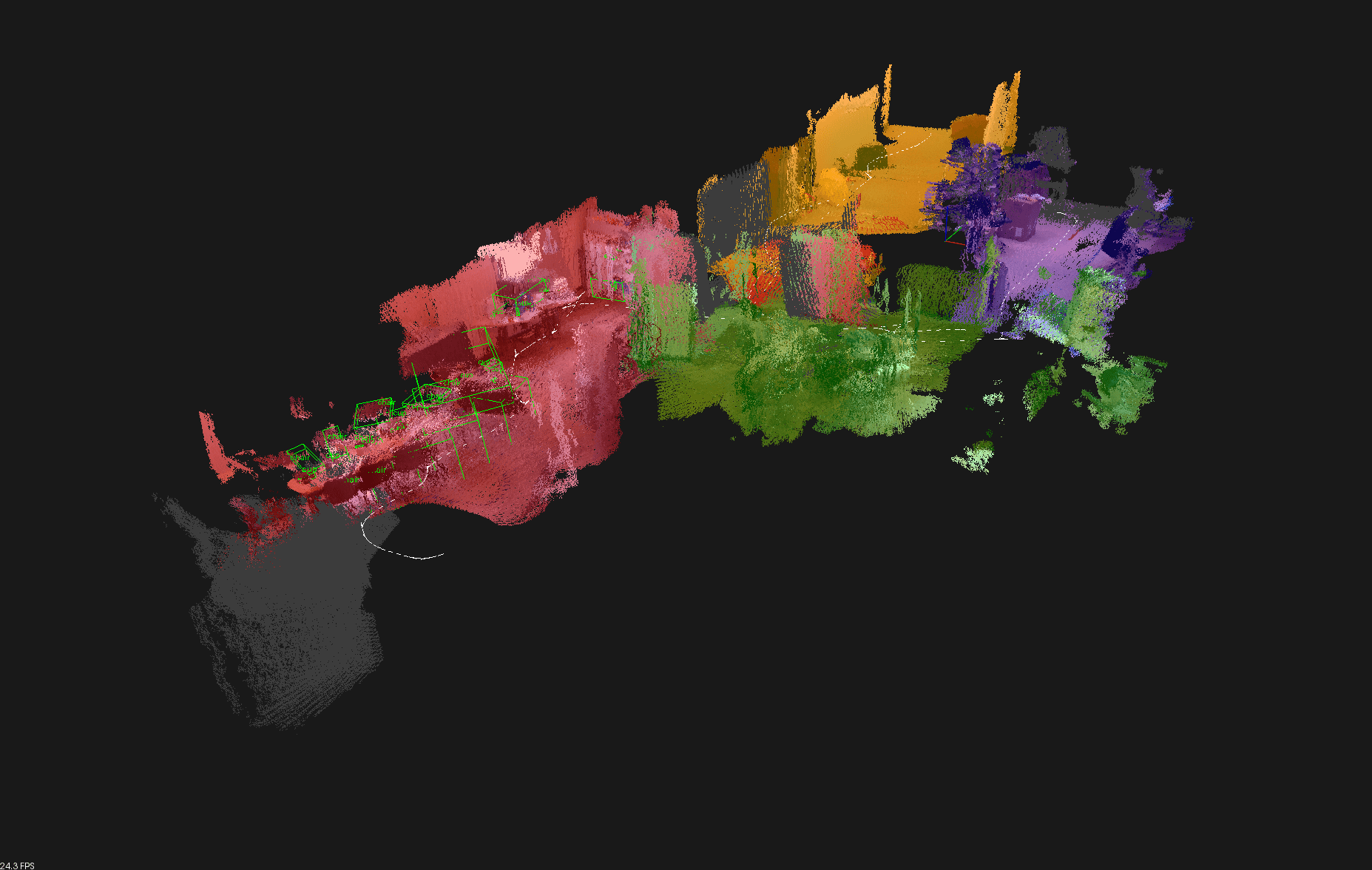}\label{fig:scene-labels-ttic}}
	\subfigure[compact environment model inferred for the command ``drive to the farthest cup in the kitchen'']{\includegraphics[width=0.325\linewidth]{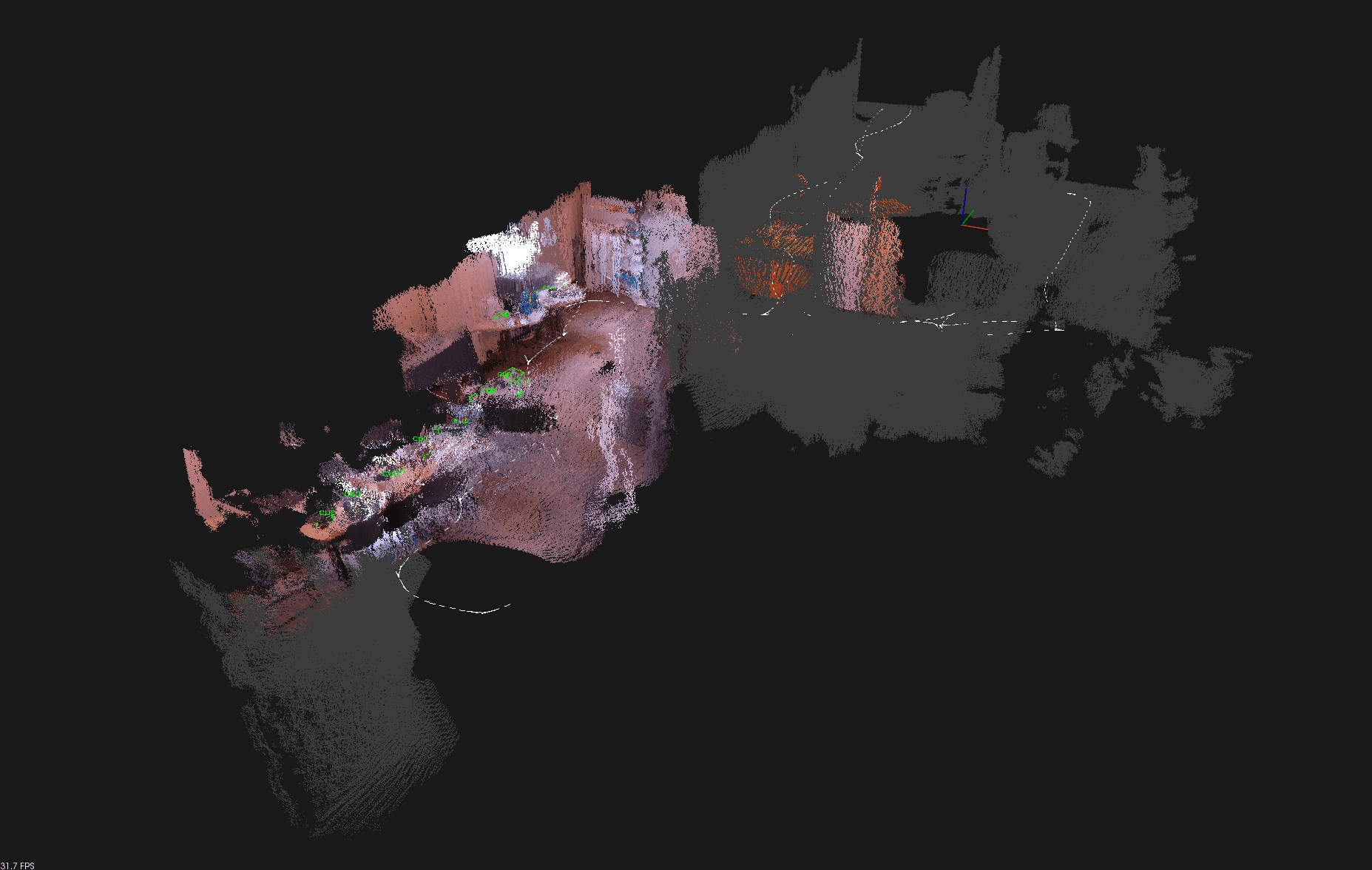}\label{fig:compact-map-ttic}}
	\subfigure[exhaustive environment model]{\includegraphics[width=0.325\linewidth]{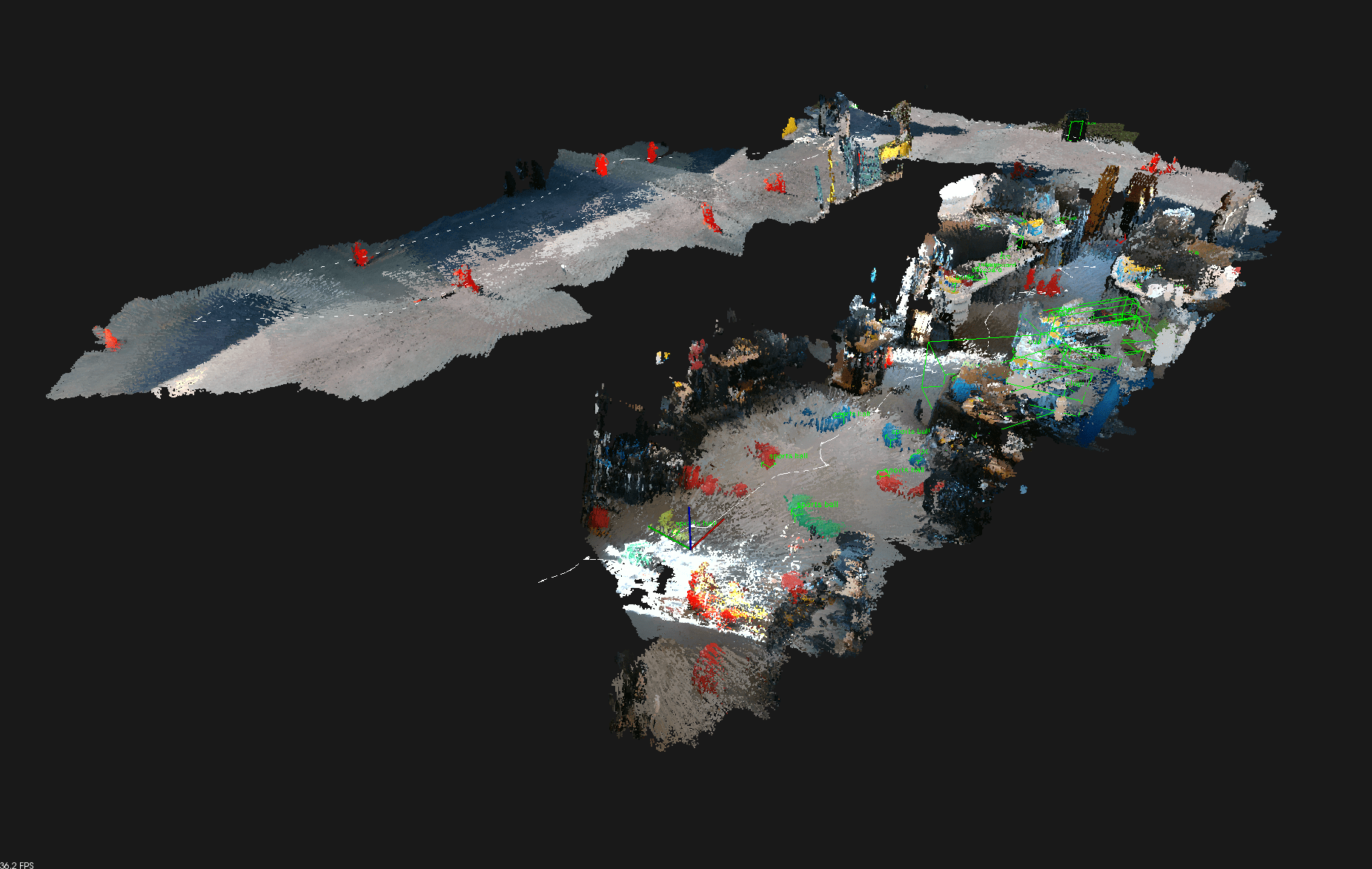}\label{fig:exhaustive-map-roc}}
	\subfigure[semantic scene labels]{\includegraphics[width=0.325\linewidth]{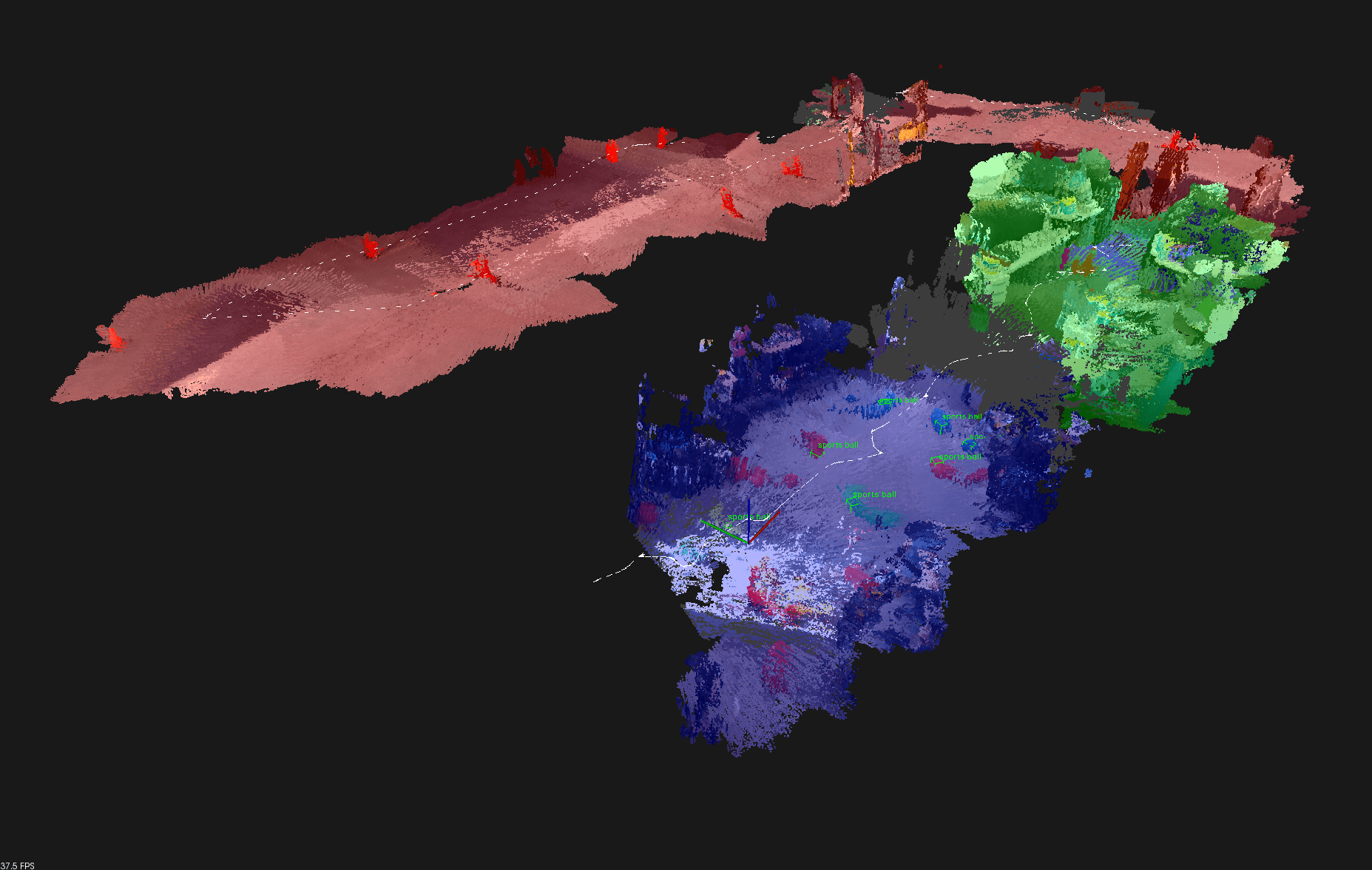}\label{fig:scene-labels-roc}}
	\subfigure[compact environment model inferred for the command ``drive to the nearest ball in the lab'']{\includegraphics[width=0.325\linewidth]{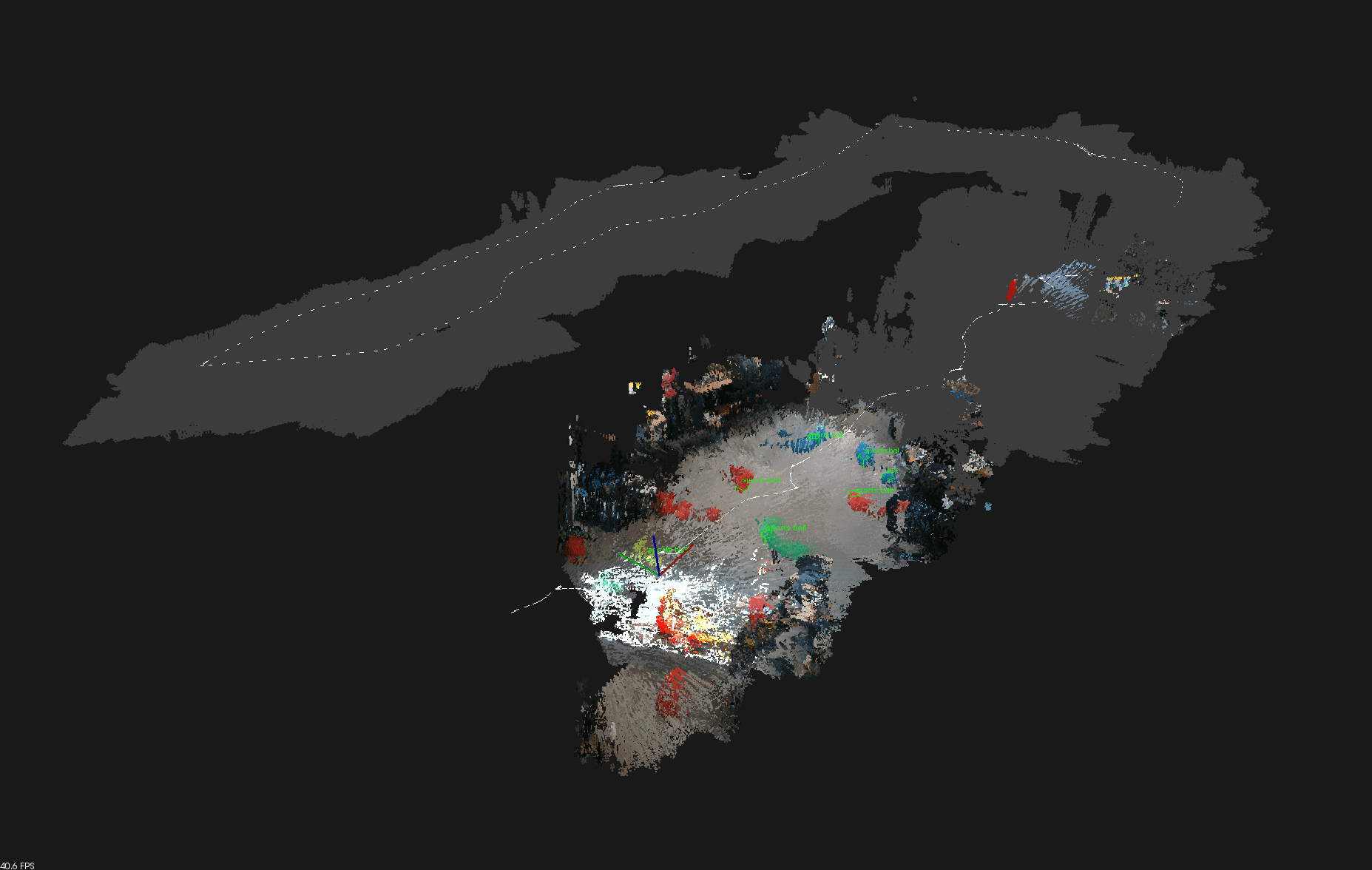}\label{fig:compact-map-roc}}\\
	\caption{A visualization of environment representations for Site 1 (top) and Site 2 (bottom). The renderings in \subref{fig:scene-labels-ttic} and \subref{fig:scene-labels-roc} depict the scene labels. The standard approach of employing all observations and object classifiers results in \subref{fig:exhaustive-map-ttic}, \subref{fig:exhaustive-map-roc} an exhaustive representation of the environment. In contrast, inferring the set of observations and detectors relevant to the command yields \subref{fig:compact-map-ttic}, \subref{fig:compact-map-roc} compact environment models that afford more efficient grounding.}
		\label{fig:observation-filtering-navigation-results}
\end{figure*}
The last model that we considered combines observation filtering with adaptive perception. The results in Table~\ref{table:results-1} show the improvement of observation filtering with adaptive perception (OF+AP) against the standard baseline (B).  As expected, combining both of these approaches reduces the time required to extract a suitable world model for natural language symbol grounding in all six scenarios. An example is depicted in Figure~\ref{fig:observation-filtering-navigation-results}.  In the best case, we observed a $90\%$ improvement in runtime performance for the instruction ``go to the nearest ball in the hallway.''  Table~\ref{table:results-2} lists the number of objects extracted by the perception pipeline. Reducing the number of objects significantly improves the runtime of symbol grounding, which is at best linear~\cite{howard14,chung15} and at worst exponential~\cite{tellex11} in the size of the world model.

\section{Conclusions} \label{sec:conclusions}

In this paper, we presented a novel framework that improves the efficiency of natural language understanding by generating and reasoning over a compact, instruction-specific world model. Underlying the framework are three primary methods that exploit the structure of language to facilitate inference.  First, we use language reduce the set of all observations available to the robot by extracting semantic labels for the context in which the salient observations occur.  Second, language is used to infer a subset of perceptual classifiers that extract a compact but sufficiently complex environment model that is suitable for interpreting the meaning of the instruction.  Third, language is used in the context of the compact environment representation to infer the symbolic meaning of the instruction. Experimental results demonstrate how adaptive perception and observation filtering improve the computational efficiency of inference without affecting the accuracy of language grounding.  In ongoing work, we are exploring methods to improve the robustness of semantic label classification for observations, including per-pixel semantic classification approaches.

This work also presents a number of interesting areas of future research.  In the examples considered here, we did not exploit prior knowledge about the environment. However, one can easily extrapolate how using past compact representations to seed future models might mitigate the need to re-classify all objects for every instruction. A model that does not discard the information, but incrementally builds a rich spatial-semantic environment model over time is likely to be highly effective and efficient for human-robot interaction in complex environments with diverse tasks.  Training and evaluating the performance of language models that use corpora collected from studies involving human-robot interaction and more complex tasks, robots, and environments that exploit differences in scale remain as future work.  Such additional experiments would further characterize the performance of the proposed model and enrich our understanding of how to best construct efficient, hierarchical representations of environments for multi-modal human-robot interaction.

\section{Acknowledgements} \label{sec:acknowledgements}
This work was supported in part by the National Science Foundation under grants IIS-1638072 and IIS-1637813, by the Robotics Consortium of the U.S. Army Research Laboratory under the Collaborative Technology Alliance Program Cooperative Agreement W911NF-10-2-0016, and by ARO grants W911NF-15-1-0402 and W911NF-17-1-0188.

\bibliographystyle{IEEEtranN}

{\small
\bibliography{references}

\begin{thebibliography}{43}
\providecommand{\natexlab}[1]{#1}
\providecommand{\url}[1]{#1}
\csname url@samestyle\endcsname
\providecommand{\newblock}{\relax}
\providecommand{\bibinfo}[2]{#2}
\providecommand{\BIBentrySTDinterwordspacing}{\spaceskip=0pt\relax}
\providecommand{\BIBentryALTinterwordstretchfactor}{4}
\providecommand{\BIBentryALTinterwordspacing}{\spaceskip=\fontdimen2\font plus
\BIBentryALTinterwordstretchfactor\fontdimen3\font minus
  \fontdimen4\font\relax}
\providecommand{\BIBforeignlanguage}[2]{{%
\expandafter\ifx\csname l@#1\endcsname\relax
\typeout{** WARNING: IEEEtranN.bst: No hyphenation pattern has been}%
\typeout{** loaded for the language `#1'. Using the pattern for}%
\typeout{** the default language instead.}%
\else
\language=\csname l@#1\endcsname
\fi
#2}}
\providecommand{\BIBdecl}{\relax}
\BIBdecl

\bibitem[Kuipers et~al.(2004)Kuipers, Modayil, Beeson, and MacMahon]{kuipers04}
B.~Kuipers, J.~Modayil, P.~Beeson, and M.~MacMahon, ``Local metrical and global
  topological maps in the {H}ybrid {S}patial {S}emantic {H}ierarchy,'' in
  \emph{Proc.\ IEEE Int'l Conf.\ on Robotics and Automation (ICRA)}, 2004.

\bibitem[Modayil et~al.(2004)Modayil, Beeson, and Kuipers]{modayil04}
J.~Modayil, P.~Beeson, and B.~Kuipers, ``Using the topological skeleton for
  scalable global metrical map-building,'' in \emph{Proc. IEEE/RSJ Int'l Conf.\
  on Intelligent Robots and Systems (IROS)}, 2004.

\bibitem[Beeson et~al.(2010)Beeson, Modayil, and Kuipers]{beeson10}
P.~Beeson, J.~Modayil, and B.~Kuipers, ``Factoring the mapping problem: {M}oble
  robot map-building in the {H}ybrid {S}patial {S}emantic {H}ierarchy,''
  \emph{Int'l J.\ of Robotics Research}, vol.~29, no.~4, 2010.

\bibitem[Pronobis and Jensfelt(2012)]{pronobis12}
A.~Pronobis and P.~Jensfelt, ``Large-scale semantic mapping and reasoning with
  heterogeneous modalities,'' in \emph{Proc.\ IEEE Int'l Conf.\ on Robotics and
  Automation (ICRA)}, 2012.

\bibitem[Walter et~al.(2013)Walter, Hemachandra, Homberg, Tellex, and
  Teller]{walter13}
M.~R. Walter, S.~Hemachandra, B.~Homberg, S.~Tellex, and S.~Teller, ``Learning
  semantic maps from natural language descriptions,'' in \emph{Proc.\ Robotics:
  Science and Systems (RSS)}, 2013.

\bibitem[Eustice et~al.(2005)Eustice, Singh, and Leonard]{eustice05}
R.~Eustice, H.~Singh, and J.~Leonard, ``Exactly sparse delayed-state filters,''
  in \emph{Proc.\ IEEE Int'l Conf.\ on Robotics and Automation (ICRA)}, 2005.

\bibitem[Olson et~al.(2006)Olson, Leonard, and Teller]{olson06}
E.~Olson, J.~Leonard, and S.~Teller, ``Fast iterative optimization of pose
  graphs with poor initial estimates,'' in \emph{Proc.\ IEEE Int'l Conf.\ on
  Robotics and Automation (ICRA)}, 2006.

\bibitem[Durrant-Whyte and Bailey(2006)]{durrant-whyte06}
H.~Durrant-Whyte and T.~Bailey, ``Simultaneous localization and mapping
  ({SLAM}): Part {I},'' \emph{IEEE Robotics and Automation Magazine}, vol.~13,
  no.~2, pp. 99--110, 2006.

\bibitem[Bailey and Durrant-Whyte(2006)]{bailey06}
T.~Bailey and H.~Durrant-Whyte, ``Simultaneous localization and mapping
  ({SLAM}): Part {II},'' \emph{IEEE Robotics and Automation Magazine}, vol.~13,
  no.~3, pp. 108--117, 2006.

\bibitem[Walter et~al.(2007)Walter, Eustice, and Leonard]{walter07}
M.~R. Walter, R.~M. Eustice, and J.~J. Leonard, ``Exactly sparse extended
  information filters for feature-based {SLAM},'' \emph{Int'l J.\ of Robotics
  Research}, vol.~26, no.~4, 2007.

\bibitem[Kaess et~al.(2008)Kaess, Ranganathan, and Dellaert]{kaess08}
M.~Kaess, A.~Ranganathan, and F.~Dellaert, ``{iSAM}: {I}ncremental smoothing
  and mapping,'' \emph{Trans.\ on Robotics}, vol.~24, no.~6, pp. 1365--1378,
  2008.

\bibitem[Cummins and Newman(2009)]{cummins09}
M.~Cummins and P.~Newman, ``Highly scalable appearance-only {SLAM} -- {FAB-MAP}
  2.0,'' in \emph{Proc.\ Robotics: Science and Systems (RSS)}, 2009.

\bibitem[Mart{\'i}nez~Mozos et~al.(2007)Mart{\'i}nez~Mozos, Triebel, Jensfelt,
  Rottmann, and Burgard]{mozos07}
O.~Mart{\'i}nez~Mozos, R.~Triebel, P.~Jensfelt, A.~Rottmann, and W.~Burgard,
  ``Supervised semantic labeling of places using information extracted from
  sensor data,'' \emph{Robotics and Autonomous Systems}, vol.~55, no.~5, 2007.

\bibitem[Zender et~al.(2008)Zender, Mart{\'\i}nez~Mozos, Jensfelt, Kruijff, and
  Burgard]{zender08}
H.~Zender, O.~Mart{\'\i}nez~Mozos, P.~Jensfelt, G.~Kruijff, and W.~Burgard,
  ``Conceptual spatial representations for indoor mobile robots,''
  \emph{Robotics and Autonomous Systems}, vol.~56, no.~6, 2008.

\bibitem[Hemachandra et~al.(2014)Hemachandra, Walter, Tellex, and
  Teller]{hemachandra14}
S.~Hemachandra, M.~R. Walter, S.~Tellex, and S.~Teller, ``Learning
  spatial-semantic representations from natural language descriptions and scene
  classifications,'' in \emph{Proc.\ IEEE Int'l Conf.\ on Robotics and
  Automation (ICRA)}, 2014.

\bibitem[Hemachandra et~al.(2015)Hemachandra, Duvallet, Howard, Roy, Stentz,
  and Walter]{hemachandra15}
S.~Hemachandra, F.~Duvallet, T.~M. Howard, N.~Roy, A.~Stentz, and M.~R. Walter,
  ``Learning models for following natural language directions in unknown
  environments,'' in \emph{Proc.\ IEEE Int'l Conf.\ on Robotics and Automation
  (ICRA)}, 2015.

\bibitem[Tellex et~al.(2011{\natexlab{a}})Tellex, Kollar, Dickerson, Walter,
  Banerjee, Teller, and Roy]{tellex11a}
S.~Tellex, T.~Kollar, S.~Dickerson, M.~Walter, A.~Banerjee, S.~Teller, and
  N.~Roy, ``Approaching the symbol grounding problem with probabilistic
  graphical models,'' \emph{AI Magazine}, vol.~32, no.~4, pp. 64--76, 2011.

\bibitem[Howard et~al.(2014)Howard, Tellex, and Roy]{howard14}
T.~M. Howard, S.~Tellex, and N.~Roy, ``A natural language planner interface for
  mobile manipulators,'' in \emph{Proc.\ IEEE Int'l Conf.\ on Robotics and
  Automation (ICRA)}, 2014.

\bibitem[Chung et~al.(2015)Chung, Propp, Walter, and Howard]{chung15}
I.~Chung, O.~Propp, M.~Walter, and T.~Howard, ``On the performance of
  hierarchical distributed correspondence graphs for efficient symbol grounding
  of robot instructions,'' in \emph{Proc. IEEE/RSJ Int'l Conf.\ on Intelligent
  Robots and Systems (IROS)}, 2015.

\bibitem[Patki and Howard(2018)]{patki18a}
S.~Patki and T.~M. Howard, ``Language-guided adaptive perception for efficient
  grounded communication with robotic manipulators in cluttered environments,''
  in \emph{Proc.\ Annual Meeting of the Special Interest Group on Discourse and
  Dialogue (SIGDIAL)}, 2018.

\bibitem[Grisetti et~al.(2009)Grisetti, Stachniss, and Burgard]{grisetti09}
G.~Grisetti, C.~Stachniss, and W.~Burgard, ``Nonlinear constraint network
  optimization for efficient map learning,'' \emph{IEEE Trans.\ on Intelligent
  Transportation Systems}, vol.~10, no.~3, 2009.

\bibitem[Kuipers(2000)]{kuipers00}
B.~Kuipers, ``The spatial semantic hierarchy,'' \emph{Artificial Intelligence},
  vol. 119, no.~1, 2000.

\bibitem[Vasudevan and Siegwart(2008)]{vasudevan08}
S.~Vasudevan and R.~Siegwart, ``Bayesian space conceptualization and place
  classification for semantic maps in mobile robotics,'' \emph{Robotics and
  Autonomous Systems}, vol.~56, no.~6, 2008.

\bibitem[Duvallet et~al.(2014)Duvallet, Walter, Howard, Hemachandra, Oh,
  Teller, Roy, and Stentz]{duvallet14}
F.~Duvallet, M.~R. Walter, T.~Howard, S.~Hemachandra, J.~Oh, S.~Teller, N.~Roy,
  and A.~Stentz, ``Inferring maps and behaviors from natural language
  instructions,'' in \emph{Proc.\ Int'l. Symp.\ on Experimental Robotics
  (ISER)}, 2014.

\bibitem[Harnad(1990)]{harnad90}
S.~Harnad, ``The symbol grounding problem,'' \emph{Physica D}, vol.~42, pp.
  335--346, 1990.

\bibitem[Tellex et~al.(2011{\natexlab{b}})Tellex, Kollar, Dickerson, Walter,
  Banerjee, Teller, and Roy]{tellex11}
S.~Tellex, T.~Kollar, S.~Dickerson, M.~R. Walter, A.~G. Banerjee, S.~Teller,
  and N.~Roy, ``Understanding natural language commands for robotic navigation
  and mobile manipulation,'' in \emph{Proc.\ Nat'l Conf.\ on Artificial
  Intelligence (AAAI)}, 2011.

\bibitem[Matuszek et~al.(2010)Matuszek, Fox, and Koscher]{matuszek10}
C.~Matuszek, D.~Fox, and K.~Koscher, ``Following directions using statistical
  machine translation,'' in \emph{Proc.\ ACM/IEEE Int'l. Conf.\ on Human-Robot
  Interaction (HRI)}, 2010.

\bibitem[Matuszek et~al.(2012)Matuszek, Herbst, Zettlemoyer, and
  Fox]{matuszek12a}
C.~Matuszek, E.~Herbst, L.~Zettlemoyer, and D.~Fox, ``Learning to parse natural
  language commands to a robot control system,'' in \emph{Proc.\ Int'l. Symp.\
  on Experimental Robotics (ISER)}, 2012.

\bibitem[Thomason et~al.(2015)Thomason, Zhang, Mooney, and Stone]{thomason15}
J.~Thomason, S.~Zhang, R.~J. Mooney, and P.~Stone, ``Learning to interpret
  natural language commands through human-robot dialog,'' in \emph{Proc.\ Int'l
  Joint Conf.\ on Artificial Intelligence (IJCAI)}, 2015.

\bibitem[Winograd(1971)]{winograd71}
T.~Winograd, ``Procedures as a representation for data in a computer program
  for understanding natural language,'' Ph.D. dissertation, Massachusetts
  Institute of Technology, 1971.

\bibitem[Roy et~al.(2003)Roy, Hsiao, and Mavridis]{roy03}
D.~Roy, K.-Y. Hsiao, and N.~Mavridis, ``Conversational robots: {B}uilding
  blocks for grounding word meaning,'' in \emph{Proc.\ HLT-NAACL Workshop on
  Learning Word Meaning from Non-Linguistic Data}, 2003.

\bibitem[Kollar et~al.(2010)Kollar, Tellex, Roy, and Roy]{kollar10}
T.~Kollar, S.~Tellex, D.~Roy, and N.~Roy, ``Toward understanding natural
  language directions,'' in \emph{Proc.\ ACM/IEEE Int'l. Conf.\ on Human-Robot
  Interaction (HRI)}, 2010.

\bibitem[Tellex et~al.(2012)Tellex, Thaker, Deits, Kollar, and Roy]{tellex12}
S.~Tellex, P.~Thaker, R.~Deits, T.~Kollar, and N.~Roy, ``Toward information
  theoretic human-robot dialog,'' in \emph{Proc.\ Robotics: Science and Systems
  (RSS)}, 2012.

\bibitem[Tellex et~al.(2014)Tellex, Knepper, Li, Rus, and Roy]{tellex14}
S.~Tellex, R.~Knepper, A.~Li, D.~Rus, and N.~Roy, ``Asking for help using
  inverse semantics,'' in \emph{Proc.\ Robotics: Science and Systems (RSS)},
  2014.

\bibitem[Gong and Zhang(2018)]{gong18}
Z.~Gong and Y.~Zhang, ``Temporal spatial inverse semantics for robots
  communicating with humans,'' in \emph{Proc.\ IEEE Int'l Conf.\ on Robotics
  and Automation (ICRA)}, 2018.

\bibitem[Spranger and Steels(2015)]{spranger15}
M.~Spranger and L.~Steels, ``Co-acquisition of syntax and semantics-an
  investigation in spatial language,'' in \emph{Proc.\ Int'l Joint Conf.\ on
  Artificial Intelligence (IJCAI)}, 2015.

\bibitem[She and Chai(2017)]{she17}
L.~She and J.~Chai, ``Interactive learning of grounded verb semantics towards
  human-robot communication,'' in \emph{Proc.\ Association for Computational
  Linguistics (ACL)}, 2017.

\bibitem[Bollini et~al.(2010)Bollini, Tellex, Thompson, Roy, and
  Rus]{bollini10}
M.~Bollini, S.~Tellex, T.~Thompson, N.~Roy, and D.~Rus, ``Interpreting and
  executing recipes with a cooking robot,'' in \emph{Proc.\ Int'l. Symp.\ on
  Experimental Robotics (ISER)}, 2010.

\bibitem[Walter et~al.(2014)Walter, Antone, Chuangsuwanich, Correa, Davis,
  Fletcher, Frazzoli, Friedman, Glass, How, Jeon, Karaman, Luders, Roy, Tellex,
  and Teller]{walter14b}
M.~Walter, M.~Antone, E.~Chuangsuwanich, A.~Correa, R.~Davis, L.~Fletcher,
  E.~Frazzoli, Y.~Friedman, J.~Glass, J.~How, J.~Jeon, S.~Karaman, B.~Luders,
  N.~Roy, S.~Tellex, and S.~Teller, ``A situationally aware voice-commandable
  robotic forklift working alongside people in unstructured outdoor
  environments,'' \emph{J.\ of Field Robotics}, 2014.

\bibitem[Paul et~al.(2016)Paul, Arkin, Roy, and Howard]{paul16a}
R.~Paul, J.~Arkin, N.~Roy, and T.~M. Howard, ``Efficient grounding of abstract
  spatial concepts for natural language interaction with robot manipulators,''
  in \emph{Proc.\ Robotics: Science and Systems (RSS)}, 2016.

\bibitem[Paul et~al.(2018)Paul, Arkin, Aksaray, Roy, and
  Howard]{paul2018efficientplatforms}
R.~Paul, J.~Arkin, D.~Aksaray, N.~Roy, and T.~M. Howard, ``Efficient grounding
  of abstract spatial concepts for natural language interaction with robot
  platforms,'' \emph{Int'l J.\ of Robotics Research}, 2018.

\bibitem[Redmon and Farhadi(2017)]{redmon2017yolo9000}
J.~Redmon and A.~Farhadi, ``Yolo9000: Better, faster, stronger,'' in
  \emph{Proc.\ IEEE Conf.\ on Computer Vision and Pattern Recognition
  (CVPR)}.\hskip 1em plus 0.5em minus 0.4em\relax IEEE, 2017.

\bibitem[Lin et~al.(2014)Lin, Maire, Belongie, Hays, Perona, Ramanan,
  Doll{\'a}r, and Zitnick]{lin2014microsoft}
T.-Y. Lin, M.~Maire, S.~Belongie, J.~Hays, P.~Perona, D.~Ramanan,
  P.~Doll{\'a}r, and C.~L. Zitnick, ``Microsoft {COCO}: Common objects in
  context,'' in \emph{Proc.\ European Conf.\ on Computer Vision (ECCV)}, 2014.

\end{thebibliography}
}

\end{document}